\documentclass[runningheads]{llncs}

\usepackage[mobile]{eccv}

\usepackage{eccvabbrv}

\usepackage{graphicx}
\usepackage{booktabs}
\usepackage[accsupp]{axessibility}  

\usepackage[breaklinks,colorlinks,citecolor=eccvblue]{hyperref}

\usepackage{orcidlink}

\begin{document}

\title{Towards High-Quality 3D Motion Transfer with Realistic Apparel Animation} 


%
\author{Rong Wang\inst{1}\orcidlink{0000-0002-1905-3175} \and
Wei Mao\inst{2}\orcidlink{0000-0002-8876-8983} \and
Changsheng Lu\inst{1}\orcidlink{0000-0002-1894-286X} \and Hongdong Li\inst{1}\orcidlink{0000-0003-4125-1554}}

\authorrunning{R.~Wang et al.}

\institute{The Australian National University \and XR Vision Labs, Tencent\\
\email{\{rong.wang, changsheng.lu, hongdong.li\}@anu.edu.au  weiwmao@global.tencent.com}}

\maketitle

\begin{center}
    \centering
    \captionsetup{type=figure}
    \includegraphics[width=0.95\textwidth]{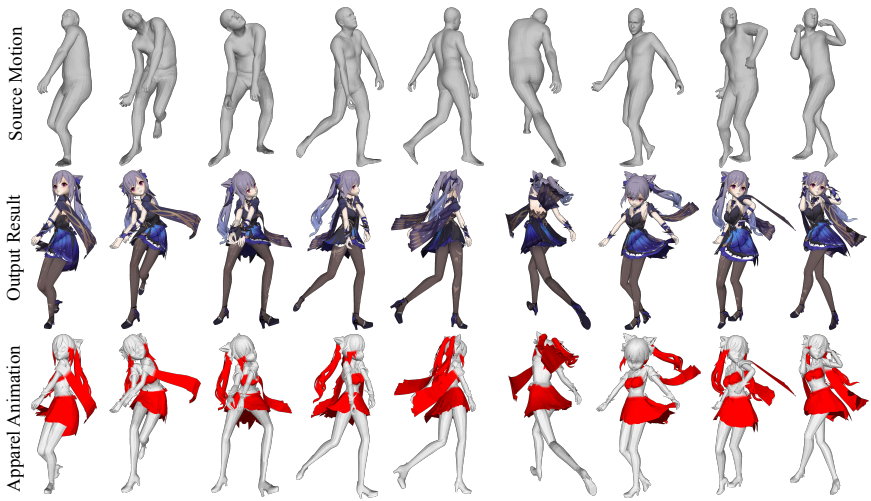}
    \captionof{figure}{We present a novel method which transfers a source motion onto a target stylized character and generates \emph{realistic apparel animation}.}
\end{center}%

\begin{abstract}
Animating stylized characters to match a reference motion sequence is a highly demanded task in film and gaming industries. Existing methods mostly focus on rigid deformations of characters' body, neglecting local deformations on the \textbf{apparel} driven by physical dynamics. They deform apparel the same way as the body, leading to results with limited details and unrealistic artifacts, \emph{e.g.} body-apparel penetration. In contrast, we present a novel method aiming for high-quality motion transfer with realistic apparel animation. As existing datasets lack annotations necessary for generating realistic apparel animations, we build a new dataset named \textbf{MMDMC}, which combines stylized characters from the \textbf{M}iku\textbf{M}iku\textbf{D}ance community with real-world \textbf{M}otion \textbf{C}apture data. We then propose a data-driven pipeline that learns to disentangle body and apparel deformations via two neural deformation modules. For body parts, we propose a geodesic attention block to effectively incorporate semantic priors into skeletal body deformation to tackle complex body shapes for stylized characters. Since apparel motion can significantly deviate from respective body joints, we propose to model apparel deformation in a non-linear vertex displacement field conditioned on its historic states. Extensive experiments show that our method produces results with superior quality for various types of apparel. Our dataset is released in \url{https://github.com/rongakowang/MMDMC}.
\end{abstract}
\section{Introduction}
3D motion transfer tackles the problem of animating a target character following a reference 3D motion sequence, \emph{e.g.} motion capture data clip. This is a long standing problem in computer vision and graphics \cite{10.5555/3285343} and is highly demanded in many applications, \emph{e.g.} digital avatars and extended reality \cite{villegas2018neural}. 

\begin{figure}[htp!]
{\includegraphics[width=\textwidth]{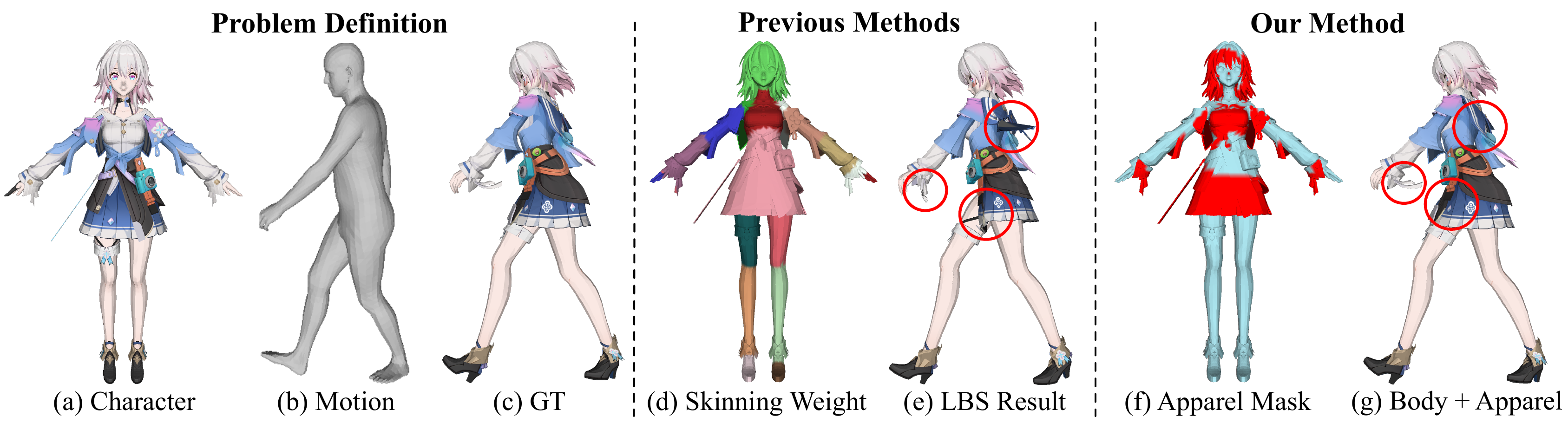}}
    \centering
    \caption{\textbf{Illustration of our method.} Given an input character (a), we aim to animate it following a reference 3D motion (b) and produce the target result (c). Previous methods mostly predict the skinning weights (d) with respect to \emph{body} joints and deform the entire character via the LBS method (e), essentially treating the apparel equally with the body. Such approach lacks visual details and often contains unrealistic artifacts, such as body-apparel penetration. In contrast, we propose a novel pipeline that discriminates apparel vertices (in \textcolor{red}{red}) by apparel segmentation (f) and then explicitly models its local deformation, thus producing realistic apparel animation (g).}
    \label{fig:open}
\end{figure}

Existing works \cite{aberman2020skeleton, li2021learning, liao2022skeleton, zhang2023skinned} for motion transfer mostly model the character deformation as solely driven by rigid transformations of body joints, where each vertex is assigned to certain body parts by the skinning weights and deformed via a skeletal deformation model, \emph{e.g.} linear blend skinning (LBS) \cite{magnenat1988joint}. However, stylized characters used in the film and game industry often feature various types of \textbf{apparel}, \emph{e.g.} garments and accessories. Such approach does not apply to apparel motions, as apparel does not have a well-defined skeleton and can \emph{locally deform additionally driven by physical dynamics}, which can be significantly different from anchored body joints. Those works neglect such local deformation on apparel and deform it the same way as the body, resulting in limited details and unrealistic artifacts, as illustrated in Figure \ref{fig:open}. 

Unfortunately, extending these works to achieve realistic apparel animation remains very challenging, primarily due to the lack of data with detailed annotations on the apparel. Existing character-motion datasets either contain minimally-clothed characters \cite{mahmood2019amass, bhatnagar2019mgn} only, or stylized characters but without proper rigging and physics simulation on the apparel \cite{Mixamo, RigNet}. In consequence, training samples from these datasets do not exhibit local apparel deformation.  

To tackle such data-insufficiency problem, in this work we first create a new dataset named \textbf{MMDMC}, which combines artist-designed characters from the \textbf{M}iku\textbf{M}iku\textbf{D}ance community \cite{MikuMikuDance} with real-world \textbf{M}otion \textbf{C}apture \cite{mahmood2019amass} data. This dataset not only features diverse and high-fidelity characters with various types of complex apparel, but is also equipped with detailed apparel annotations, including rigging, segmentation and physics simulation designed by professional artists, therefore makes it amenable to the learning of apparel animation.

Leveraging the rich data, we then develop a novel method for high-quality motion transfer, which can notably generate realistic and vivid apparel animation. Specifically, our model learns to discriminate and separately generate body and apparel deformation. For body parts, we learn to predict the skinning weights and deform them via the skeletal deformation method, utilizing a novel geodesic attention block to tackle the complex body shapes of stylized characters and effectively incorporate semantic priors in body deformation. For apparel, we model its deformation in a non-linear per-vertex displacement field conditioned on its historic states, which allows us to generate local apparel deformation independent to the motion of body joints. Finally, we jointly refine results from both modules to encourage continuity as well as penalize body-apparel penetration.

Our contributions can be summarized as follows. \emph{(i)}
We introduce a new dataset MMDMC, which features diverse and complex character apparel with detailed annotations of rigging and physics simulation (Section \ref{sec_data}). \emph{(ii)} We propose a novel method for high-quality 3D motion transfer with apparel animation generation, which learns to effectively disentangle body and apparel deformation (Section \ref{sec_method}). Extensive experiments show that our method produces superior results on various types of motion and apparel.

\section{Related Works}
\textbf{3D Motion Transfer. } 
Since human motions can be described by rigid transformations of articulated body joints, existing works \cite{villegas2018neural, aberman2020skeleton, li2021learning, zhang2023skinned, chen2023weakly} on 3D motion transfer often assume a known skeleton template, which can be obtained from statistical models like SMPL \cite{SMPL:2015} or enveloped by neural rigging methods \cite{AnimSkelVolNet, RigNet}. Such a skeleton-based approach typically learns to estimate the vertex skinning weights and deform the character via a skeletal deformation model, such as linear blend skinning (LBS). In particular, \cite{aberman2020skeleton} proposes a skeleton-aware network to tackle the challenge of transferring motion between skeletons with topologically different connections. \cite{li2021learning} further mitigates the deformation artifacts in the LBS method with residual neural blend shapes.
\cite{villegas2021contact, zhang2023skinned} assume skinned characters and adopts geometry priors to refine body contact and collision. While they achieve promising results on body deformation, it is difficult to extend their works to deform apparel since there is no a unified apparel skeleton. Moreover, the apparel like loose garments can deform largely under physical dynamics and do not closely follow the motion of body joints. Hence, deforming apparel based on skinning weights to \emph{body joints} often produces undesired discontinuity \cite{zhao2023learning}. Several works \cite{zheng2021deep, yifan2020neural} explore to incorporate non-rigid deformation in motion transfer, however, they assume simplified deformation models thus presenting limited diversity and fidelity.

Alternatively, \cite{liao2022skeleton, wang2023hmc, wang2023zero} propose to adopt a skeleton-free approach to mitigate the restrictions of pre-defined skeletons.
In particular, \cite{liao2022skeleton} pioneers the work by estimating the transformations for joints that are not necessarily articulated, therefore can flexibly handle arbitrary skeleton structures. \cite{wang2023hmc} further extends this work with a hierarchical mesh coarsening strategy to better preserve motion semantics in low-resolution meshes. While \cite{liao2022skeleton} can introduce virtual joints on apparel in principle, consistently estimating skinning weights on complex apparel remains challenging, and the dynamical effects are difficult to recover. Meanwhile, \cite{wang2023zero} proposes to predict dense vertex displacements in an implicit neural deformation module, by pre-training on minimally-clothed human meshes from the AMASS \cite{mahmood2019amass} motion dataset. However, this work identifies apparel as an extension to body parts, and does not distinguish apparel deformation. \\

\noindent
\textbf{Apparel Animation Generation. }Generating realistic animation for apparel (or only the garments) conditioned on the body motion and underlying physical dynamics has been widely studied in related works \cite{ wang2019learning, patel2020tailornet, zhang2022motion, pan2022predicting, zhao2023learning}. Specifically, \cite{wang2019learning} proposes a garment generative model to learn intrinsic physics properties that determine the garment deformation. \cite{patel2020tailornet} decouples high-frequency components in garment deformation to model wrinkle effects. \cite{zhang2022motion} leverages temporal information to learn time-dependent dynamic skinning weights for garment vertices. \cite{pan2022predicting, zhao2023learning} proposes to model loose garment motion as driven by learned virtual anchors, which mimics the physics simulation process. While all above methods learn the deformation for a single garment, \cite{shao2023towards} further extends to predict for multiple layered garments.
However, all these methods assume known garment templates thus do not apply on a holistic character mesh, \emph{e.g.} results generated by recent AIGC works \cite{cao2023dreamavatar,canfes2023text}, and extracting separate garment layers often requires time-consuming registration and optimization\cite{xiang2021modeling, xiang2022dressing}. More importantly, they assume known body shapes, \emph{e.g.} a SMPL \cite{SMPL:2015} template, and do not estimate body motions, therefore can not apply to motion transfer for \emph{stylized} characters, which requires estimating deformations for complex body shapes. In contrast to these works, we tackle both body and apparel deformation with explicit segmentation of apparel components, which is far more challenging. In following sections, we will introduce the data and method of our work.

\section{The MMDMC Dataset}
\label{sec_data}
Existing paired character-motion datasets contain either minimally-clothed characters only \cite{mahmood2019amass, bhatnagar2019mgn}, or stylized characters but lack rigging and physics simulation for apparel \cite{RigNet, Mixamo}. The apparel on their data can not deform locally and therefore is not suitable for the purpose of our work. To tackle this issue, we introduce the \textbf{MMDMC} dataset, which is the first dataset for motion transfer with detailed apparel annotations. We compare with other datasets in Table \ref{table:data}.

\begin{table}[htp!]
\centering
  \caption{\textbf{Comparison of datasets.} Existing publicly available datasets \cite{mahmood2019amass, bhatnagar2019mgn, RigNet, Mixamo} cannot facilitate the training of apparel deformation due to the lack of proper apparel annotations. In contrast, the MMDMC dataset features rigs, segmentation and physics simulation for the apparel in a large number of character-motion sample pairs.} 
  \resizebox{0.85\linewidth}{!}{
  \setlength{\tabcolsep}{3pt}
  \begin{tabular}{lccccc}
    \toprule
    \multicolumn{1}{l}{Datasets} & \#Character & \#Motion (min.) & Apparel & Apparel Rigs &  Apparel Physics      \\ \hline
    \centering
  AMASS \cite{mahmood2019amass} & 344 & 2420.9 & No & No & No \\
  MGN \cite{bhatnagar2019mgn} & 96 & - & Yes\footnotemark[2] & No & No \\
  RigNetv1 \cite{RigNet} & 620\footnotemark[1] & - & Yes & No & No\\
  Mixamo \cite{Mixamo} & 118 & 124.2 & Yes & No & No \\ \hline
  MMDMC & 102 & 203.4 & Yes & \textbf{Yes} & \textbf{Yes} \\ 
  \bottomrule
  \end{tabular}}
  \label{table:data}
\end{table}

\noindent
\textbf{Characters. }We collect 125 publicly available characters from two games: \emph{Genshin Impact} and \emph{Honkai: Star Rail}, which contain diverse and complex apparel. Apparel are first identified by professional artists that are considered locally deformable, \emph{e.g.} loose garments and long hairs. To improve simulation efficiency, only components that have \emph{notable} physical dynamics are marked as apparel, \emph{e.g.} lower-half skirt of a one-piece garment. We follow such common design in game industries when annotating ground truth apparel vertices. All apparel parts are then manually rigged and set with physical parameters using the MMD \cite{MikuMikuDance} software. To ensure compatibility with the mocap data, we adjust the body skeleton to align with the SMPLH \cite{romero2022embodied} and Mixamo \cite{Mixamo} models. We show and compare the rigging annotations on sample characters in Figure \ref{fig:data}. More details for the annotation process are included in the supplementary materials.

\setcounter{footnote}{1} 
\footnotetext{We count only humanoid characters in RigNetv1.}
\setcounter{footnote}{2} 
\footnotetext{Mostly minimally-clothed with limited types of apparel.}

\begin{figure}[htp!]
{\includegraphics[width=\textwidth]{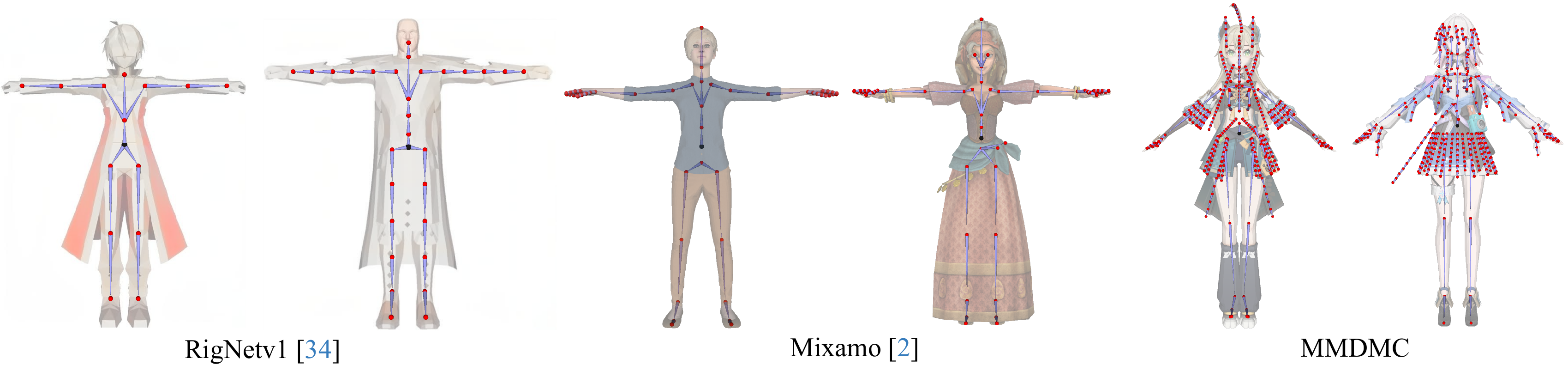}}
    \centering
    \caption{\textbf{Comparison of rig annotations with existing datasets.} Existing datasets \cite{RigNet, Mixamo} mostly provide rigging on body parts, while the apparel is not rigged in detail, hence the apparel can not independently deform. In contrast, our dataset contains dense apparel rigs, thus enabling realistic ground truth apparel animation.}
    \label{fig:data}
\end{figure}

\noindent
\textbf{Motion. }To enforce realism and diversity of the reference motion, we select 120 motion sequences from the AMASS \cite{mahmood2019amass} motion dataset, which contains motion capture data collected from human actors. Since our characters are fully rigged and skinned, we directly apply joint rotations and translations onto the characters' body skeleton and render the results in Blender \cite{blender}. The  apparel animations are simulated by the Bullet \cite{coumans2021} physics simulator with manually set physical parameters as introduced above. In particular, we enforce all motions to start from the rest pose and follow \cite{patel2020tailornet} to relax the characters for the first few frames to avoid the physical artifacts caused by the sudden pose change. Leveraging rich data, we then propose a novel data-driven pipeline for motion transfer with realistic apparel animation generation.

\begin{figure*}[htp!]
{\includegraphics[width=0.95\textwidth]{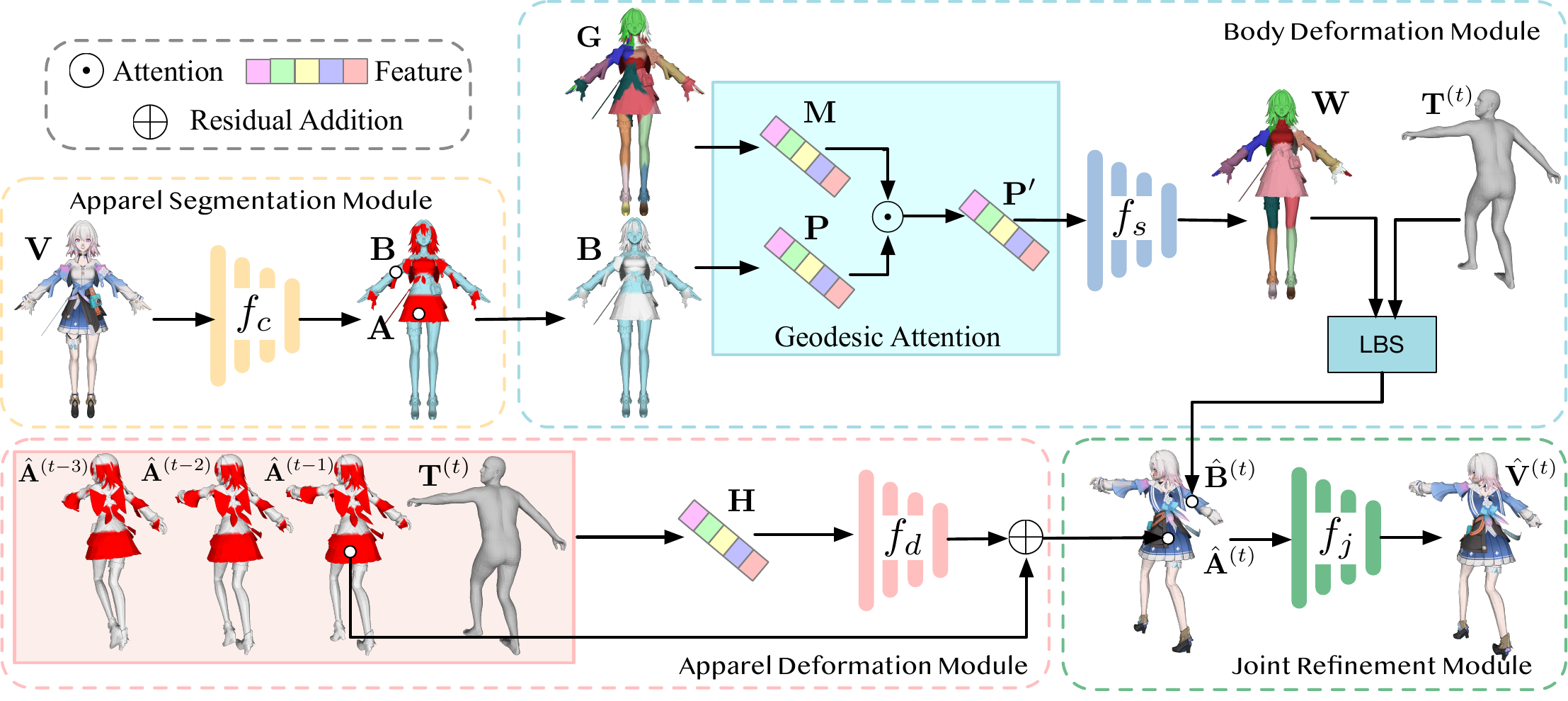}}
    \centering    \caption{\textbf{Overview of our method.} Given the input character $\mathbf{V}$ (with known joint positions), our model first discriminates body ($\mathbf{B}$, blue) and apparel ($\mathbf{A}$, red) vertices in an apparel segmentation module. With the reference joints motion $\mathbf{T}^{(t)}$, we propose a geodesic attention block to estimate the skinning weight $\mathbf{W}$ and deform the body via the LBS method. Moreover, we model non-linear apparel displacement conditioned on historic states and joint motions. Finally, we jointly refine outputs from both modules to obtain the overall result $\hat{\mathbf{V}}^{(t)}$.}
    \label{fig:main}
\end{figure*}

\section{Method}
\label{sec_method}

\textbf{Problem Definition.} Given an input character mesh of $N$ vertices $\mathbf{V} \in \mathbb{R}^{N \times 3}$ and $J$ joints $\mathbf{J} \in \mathbb{R}^{J \times 3}$, we aim at deforming it to generate an animation of $T$ meshes $\{\mathbf{\hat{V}}^{(t)}\}_{t = 1}^T$, which follows a reference motion sequence $\{\mathbf{T}^{(t)}\}_{t = 1}^T$. In particular, $\mathbf{T}^{(t)} \in \mathbb{R}^{J \times 3 \times 4}$ represents $J$ joint transformations at time $t$, namely $\mathbf{T}^{(t)}_j = [\mathbf{R}^{(t)}_j, \mathbf{t}^{(t)}_j]$ contains the rotation $\mathbf{R}^{(t)}_j\in\mathbb{R}^{3\times 3}$ and translation $\mathbf{t}^{(t)}_j\in\mathbb{R}^3$ of the $j$-th joint. We follow \cite{li2021learning,zhang2023skinned} to adopt this motion representation as it is compatible with the mocap data format and simplifies relevant applications \cite{li2021learning}. Note that we do not assume input skeletons for apparel as a unified skeleton for all apparel does not exists.
Formally, we aim to learn a general function $\mathcal{F}(\cdot)$ parameterized by deep networks such that:
\begin{equation}
    \mathcal{F}(\mathbf{V}, \mathbf{J}, \{\mathbf{{T}}^{(t)}\}_{t = 1}^T) = \{\mathbf{\hat{V}}^{(t)}\}_{t = 1}^T \;.
\end{equation}

\noindent
\textbf{Overview. }As shown in Figure \ref{fig:main}, we propose a data-driven pipeline to separately deform body and apparel, since they comply with different deformation constraints in principle. Specifically, we first train a binary classifier $f_c(\cdot)$ in an apparel segmentation module to discriminate apparel vertices $\mathbf{A}\in\mathbb{R}^{N_a \times 3}$ from body vertices $\mathbf{B}\in\mathbb{R}^{N_b \times 3}$ on the input character mesh (Section \ref{sec_main0}), where $N_a+N_b = N$. For body vertices $\mathbf{B}$, we learn a skinning weight predictor $f_s(\cdot)$ to predict the skinning weights $\mathbf{W} \in \mathbb{R}^{N_b \times J}$ and then generate the body deformation $\hat{\mathbf{B}}^{(t)}$ frame-by-frame via the LBS method (Section \ref{sec_main1}). For apparel vertices $\mathbf{A}$, we learn a residual displacement network $f_d(\cdot)$ to generate non-linear apparel deformation $\hat{\mathbf{A}}^{(t)}$ conditioned on temporal clues of historic apparel states $\{\hat{\mathbf{A}}^{(t-k)}\}_{k=1}^{K=3}$ (Section \ref{sec_main2}) in a clip of length $T$. Finally, we jointly refine results from both modules to obtain the result $\hat{\mathbf{V}}^{(t)}$ (Section \ref{sec_main3}).

\subsection{Apparel Segmentation Module}
\label{sec_main0}
Given an input character mesh $\mathbf{V}$, we wish to identify apparel parts that can locally deform independent to body parts, which we formulate as a binary classification problem. To this end, we train a classifier $f_c(\cdot)$ that contains a PointNet~\cite{qi2017pointnet}-based encoder to extract geometric features from character vertex positions followed by an MLP decoder to predict the vector $\hat{\mathbf{p}} \in \mathbb{R}^N$, which represents the probability of each vertex being as apparel:
\begin{equation}
     \hat{\mathbf{p}} = \text{Sigmoid}(f_c(\mathbf{V})) \;.
\end{equation}

We train this classifier using a binary cross entropy loss with the ground truth apparel mask $\bar{\mathbf{p}}$ as:
\begin{equation}
    \mathcal{L}_b = \frac{1}{N}\sum_{n=1}^N\hat{\mathbf{p}}_n \text{log} (\bar{\mathbf{p}}_n) +(1 - \hat{\mathbf{p}}_n) \text{log}(1 -\bar{\mathbf{p}}_n) \;.
\end{equation}

With the trained apparel segmentation module, we obtain the partition of apparel vertices $\mathbf{A}$ and body vertices $\mathbf{B}$ from the input character, which is the foundation for separate body and apparel deformation.

\subsection{Body Deformation Module}
\label{sec_main1}
Given the reference motion for $J$ body joints $\mathbf{J}$ at time frame $t$, the body deformation can be efficiently computed via the LBS \cite{magnenat1988joint} method as: \begin{equation}
    \hat{\mathbf{B}}_{i}^{(t)} = \sum_{j=1}^J \mathbf{W}_{ij} \cdot (\mathbf{R}_j^{(t)} ( \mathbf{B}_{i} - \mathbf{J}_j) + \Tilde{\mathbf{t}}_j^{(t)}) \;,
    \label{eq_lbs}
\end{equation}

where $\mathbf{B}_{i}^{(t)} \in \mathbb{R}^3$ represents the $i$-th body vertex, $\hat{\mathbf{B}}_{i}^{(t)}$ represents its deformed position, and $\Tilde{\mathbf{t}}_j^{(t)}$ represents the translation scaled by the bone lengths of the input character. We therefore aim to learn the skinning weight $\mathbf{W}$ in a neural predictor $f_s(\cdot)$. Specifically, we first extract per-joint features $\mathbf{P} \in \mathbb{R}^{N_b \times J \times D}$ for each body vertex in a MLP encoder, which captures the spatial relationship between the vertex and each joint in a $\mathbb{R}^{D}$ vector. Since a body vertex can only be influenced by a limited set of joints \cite{SMPL:2015}, we adopt a geodesic attention block to adaptively aggregate features from notable joints and suppress features from irrelevant joints. Motivated by \cite{RigNet}, we compute the vertex-joint geodesic distance matrix $\mathbf{G} \in \mathbb{R}^{N_b \times J}$, where $\mathbf{G}_{ij}$ represents the geodesic distance, \emph{i.e.} shortest path length along the mesh edges, between the vertex $\mathbf{v}_i$ and joint $\mathbf{J}_j$\footnote{Since joints are often defined inside the mesh but not on the surface, we associate each joint with its closest vertex on the mesh surface as an anchor vertex for geodesic distance computation.}. $\mathbf{G}_i$ serves as a semantic prior for aggregating features of $i$-th vertex, as this vertex can be roughly segmented to the joint that has the minimal distance. Note that \cite{RigNet} directly selects top joint features based on \emph{raw} geodesic distances, which are often noisy for vertices close to several joints, as shown in Figure \ref{fig:ab1}. Alternatively, we convert $\mathbf{G}$ into a learnable attention \cite{vaswani2017attention} map $\mathbf{M}$ as:
\begin{equation}
    \mathbf{M} = \text{Softmax}(f_m(\mathbf{G})) \;,
\end{equation}

where $\mathbf{M} \in \mathbb{R}^{N_b \times J}$ consists of the weights for each joint feature, and $f_m(\cdot)$ is an MLP that encodes the raw geodesic distance. We then weight per-joint features by $\mathbf{M}$ and obtain the fused feature map $\mathbf{P}' \in \mathbb{R}^{N_b \times D}$, where:
\begin{equation}
  \mathbf{P}'_{i} = \sum_j \mathbf{P}_{ij} \cdot \mathbf{M}_{ij} \;.
\end{equation}

 Finally, we use another PointNet-based network $f_s(\cdot)$ to predict the skinning weight as:
\begin{equation}
  \mathbf{W} = \text{Softmax}(f_s(\mathbf{P}')) \;.
\end{equation}

\noindent
\textbf{Training Losses. }We follow \cite{li2021learning} to indirectly supervise the skinning weight prediction by measuring the L1 distance between deformed and ground truth mesh $\bar{\mathbf{B}}^{(t)}$ as:
\begin{equation}
    \mathcal{L}_{vb} = \frac{1}{N_b}\sum_{i=1}^{N_b}\| \hat{\mathbf{B}}_i^{(t)} - \bar{\mathbf{B}}_i^{(t)} \|_1 \;.
\end{equation}

Besides, we follow \cite{liao2022skeleton} to impose an edge loss to penalize flying vertices and irregular surface as:
\begin{equation}
    \mathcal{L}_{eb} =\frac{1}{|\mathcal{E}_b|} \sum_{\{i, j\} \in \mathcal{E}_b}\left\lvert\|\hat{\mathbf{B}}_{i}^{(t)} - \hat{\mathbf{B}}_{j}^{(t)}\|_2 - \|{\mathbf{B}}_{i} - {\mathbf{B}}_{j}\|_2\right\rvert
\end{equation}

where $\mathcal{E}_b$ represents the set of edges connecting body vertices, and $|\mathcal{E}_b|$ is the total number of body edges. Furthermore, we regularize the skinning weights by imposing a smoothness loss to enforce continuity as:
\begin{equation}
    \mathcal{L}_s = \frac{1}{|\mathcal{E}_b|} \sum_{\{i, j\} \in \mathcal{E}_b}\|\mathbf{W}_i - \mathbf{W}_j\|_2\;,
\end{equation}

where $\mathbf{W}_i\in\mathbb{R}^{J}$ is the skinning weights for the $i$-th vertex.
Finally, the overall training objectives for the body deformation module is a weighted sum of indivudal loss as $
\mathcal{L}_b = \lambda_{vb}\mathcal{L}_{vb} + \lambda_{eb}\mathcal{L}_{eb} + \lambda_s\mathcal{L}_s$.

\subsection{Apparel Deformation Module}
\label{sec_main2}
Unlike body parts, apparel such as loose-fitting garments can deform largely under the effects of physical dynamics and do
not closely follow the motion of body joints. Hence using the skinning weights with respect to body joints to deform apparel often generates discontinuity and undesired artifacts, such as body-apparel penetration. To address this issue, we propose to model residual apparel deformation as non-linear \emph{per-vertex displacement field}, conditioned on the reference motion and historic apparel states as:
\begin{align}
  \Delta \hat{\mathbf{A}}^{(t)} = f_d(\{\hat{\mathbf{A}}^{(t-k)}\}_{k=1}^{K=3},\mathbf{T}^{(t)})\;, \quad\quad  
  \hat{\mathbf{A}}^{(t)} = \hat{\mathbf{A}}^{(t-1)} + \Delta \hat{\mathbf{A}}^{(t)} \;.
\label{eq3}
\end{align}

Specifically, given the segmented apparel graph of apparel vertices and edges, we first extract apparel features $\mathbf{H} \in \mathbb{R}^{N_a \times (9 + K)}$ for each node as:
\begin{equation}
    \mathbf{H} = \mathbf{A}^{(t-1)} \oplus \dot{\mathbf{A}}^{(t-1)} \oplus \ddot{\mathbf{A}}^{(t-1)} \oplus f_t(\mathbf{T}^{(t)})\;,
\end{equation}

where $\dot{\mathbf{A}}^{(t-1)}$ and $\ddot{\mathbf{A}}^{(t-1)}$ are the discrete apparel velocity and acceleration, $\oplus$ represents channel-wise concatenation and $f_t(\cdot)$ is a MLP encoder that encodes body motion into a $\mathbb{R}^K$ global feature. This apparel feature $\mathbf{H}$ contains both local features of historic apparel states and a global feature of joint motion, which can effectively guide the learning of apparel deformation. We then refine apparel features using GCN with edge-convolution blocks \cite{wang2019dynamic} as:
\begin{equation}
    \mathbf{H}_{i}' = \max_{j \in \mathcal{N}(i)}\text{MLP}(\mathbf{H}_{i}, \mathbf{H}_{j} - \mathbf{H}_{i}) \;,
\end{equation}

where $\mathcal{N}(i)$ is the neighbour index set for the $i$-th node on the apparel graph. Note that different components of character apparel, \emph{e.g.} hair and garments, will individually form disconnected sub-graphs, hence the edge-convolution is \emph{grouped} and features from different components will not corrupt with each other, which allows us to simultaneously learn deformations of all apparel components. Finally, we forward the refined apparel feature from the last edge-convolution block to a MLP decoder to produce $\Delta \hat{\mathbf{A}}^{(t)}$ as described in Eq.(\ref{eq3}).

\noindent
\textbf{Training Losses.}
Since the MMDMC dataset provides realistic ground truth apparel animation, we directly supervise the apparel displacement prediction as:
\begin{equation}\label{eq1}
    \mathcal{L}_{va} = \| \hat{\mathbf{A}}^{(t)} - \bar{\mathbf{A}}^{(t)} \|_1 \;,
\end{equation}

where $\bar{\mathbf{A}}^{(t)}$ represents the ground truth apparel vertex positions. We also impose an edge loss as:
\begin{equation}
\label{eq2}
    \hspace{-0.2cm}\mathcal{L}_{ea} = \frac{1}{|\mathcal{E}_a|}\sum_{\{i, j\} \in \mathcal{E}_a}\left\lvert\|\hat{\mathbf{A}}_{i}^{(t)} - \hat{\mathbf{A}}_{j}^{(t)}\|_2 - \|{\mathbf{A}}_{i} - {\mathbf{A}}_{j}\|_2\right\rvert
\end{equation}

where $\mathcal{E}_a$ is the set of edges in the apparel graph. Finally, the overall training objectives for the apparel deformation module is $
    \mathcal{L}_a = \lambda_{va}\mathcal{L}_{va} + \lambda_{ea}\mathcal{L}_{ea}$.

\subsection{Joint Refinement Module}
\label{sec_main3}
Since the apparel and body are separately deformed via distinct approaches, directly tiling outputs from individual module can cause discontinuity at the body-apparel boundary. In addition, imperfect apparel segmentation may cause the apparel to falsely deformed by the body module. To address these issues, we use a light-weight MLP as a joint refinement network $f_j(\cdot)$ to refine and merge results from both modules as:
\begin{align}
\Delta \hat{\mathbf{V}}^{(t)} = f_j(\hat{\mathbf{A}}^{(t)},  \hat{\mathbf{B}}^{(t)})\;, \quad\quad \hat{\mathbf{V}}^{(t)} = \text{tile}(\hat{\mathbf{A}}^{(t)},  \hat{\mathbf{B}}^{(t)}) + \Delta \hat{\mathbf{V}}^{(t)} \;,
\end{align}

where $\hat{\mathbf{A}}^{(t)}, \hat{\mathbf{B}}^{(t)}$ are tiled in the same order as the input character mesh. We train $f_j(\cdot)$ with similar vertex and edge losses $\mathcal{L}_v$ and $\mathcal{L}_e$ respectively as in Eq.(\ref{eq1}) and Eq.(\ref{eq2}), but use ground truth vertex position $\bar{\mathbf{V}} \in \mathbb{R}^{N \times 3}$ and edges $\mathcal{E}$ from the entire character mesh. We observe the edge loss between body and apparel also facilitate to mitigate potential drifting issues in apparel deformation. In addition, we impose a L2 regularization loss $\mathcal{L}_r$ on $\Delta\hat{\mathbf{V}}^{(t)}$ to mitigate artifacts introduced in per-vertex body displacements. Finally, the overall training objectives for the joint refinement network is $
    \mathcal{L}_j = \lambda_{v}\mathcal{L}_{v} + \lambda_{e}\mathcal{L}_{e} + \lambda_{r}\mathcal{L}_{r}$.

\section{Experiments}
\subsection{Datasets}
We evaluate our method on two datasets: MMDMC and Mixamo \cite{Mixamo}. We randomly select 5 characters and test on unseen motion clips for evaluation. Since the Mixamo dataset does not provide ground truth segmentation, rigging and physics properties for apparel, it is not suitable for quantitative comparison. Alternatively, we randomly select clothed humanoid characters to qualitatively evaluate the generalizability of our method. We include the character license used in the MMDMC dataset in the supplementary materials.

\begin{figure*}[htp!]
{\includegraphics[width=0.95\textwidth]{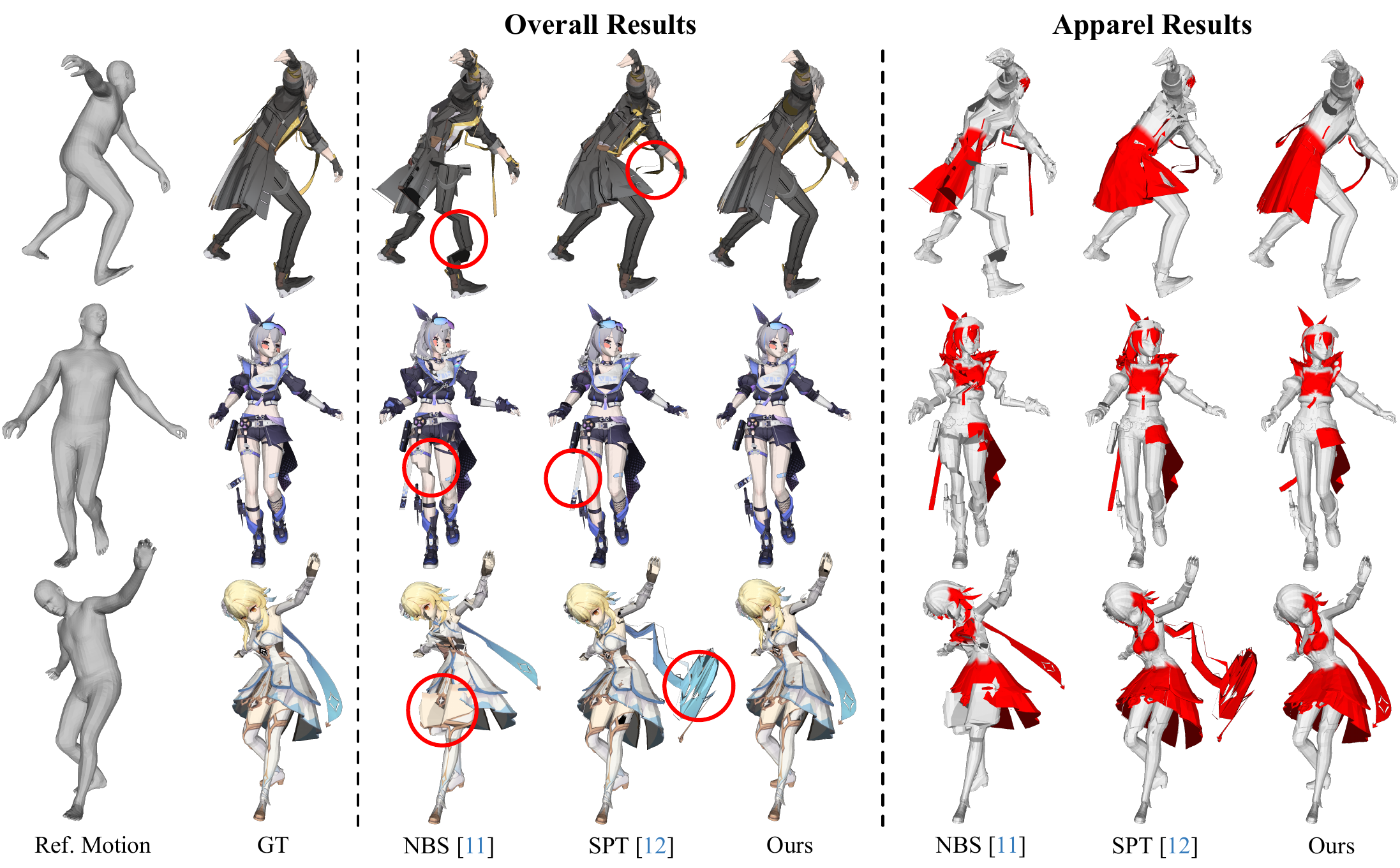}}
    \centering    
    \caption{\textbf{Qualitative comparison.} Our method produce superior results than baseline methods \cite{li2021learning, liao2022skeleton} that both contain artifacts on body or apparel (in circles). Moreover, we generate more realistic apparel results as highlighted in red at the right (using the GT apparel mask for consistent visualization of baseline methods).}
    \label{fig:quali}
\end{figure*}

\subsection{Implementation Details}
We implement all modules in PyTorch \cite{paszke2019pytorch}, and train them using the AdamW \cite{loshchilov2017decoupled} optimizer with a learning rate of $1 \times 10^{-4}$. All models are trained on a single NVIDIA RTX 3090 GPU, where the apparel segmentation and body deformation modules are trained independently and rest modules are trained end-to-end. We compute the geodesic matrix for each input character mesh using the NeworkX \cite{hagberg2008exploring}, and use $J = 40$ body joints from the SMPLH \cite{romero2022embodied} model for both reference motions and target characters. In each motion clip, We initialise apparel states from the ground truth and set the clip length as $T = 10$ frames when training and testing the apparel deformation module due to limited computation resource. In consequence, the results can still drift for testing sequences with significantly larger length.
For the hyper-parameters, we follow \cite{liao2022skeleton} to set $\lambda_{v*}$ and $\lambda_{e*}$ to be 1.0 and 100 for all vertex and edge losses, and $\lambda_{s}, \lambda_{r}$ to be 0.01.

\subsection{Metrics and Baselines}
\textbf{Metrics.} We follow \cite{wang2023zero, liao2022skeleton} to evaluate Point-wise Mesh Euclidean Distance
(PMD) \cite{wang2020neural}, which measures the average Euclidean distance between the predicted and ground truth deformed mesh. Since baseline methods do not explicitly estimate apparel deformation, we separately report PMD$_a$ on apparel vertices and PMD$_b$ on body vertices respectively, partitioned by the ground truth apparel mask.
In addition, we follow \cite{wang2023zero} to report the edge length
score (ELS) to evaluate the smoothness of the deformation.

\noindent
\textbf{Baselines. }We follow \cite{wang2023zero} to compare our method with two recent works: Neural Blend Shapes (NBS) \cite{li2021learning} and Skeleton-free Pose Transfer (SPT) \cite{liao2022skeleton}. NBS is a skeleton-based method that deforms apparel via the LBS method, but using the skinning weights with respect to body joints. SPT is a skeleton-free method that can flexibly infer joint positions, potentially including virtual joints on the apparel. We implement both methods using the official code and train on the MMDMC dataset with the same setting as ours. For a fair comparison, we provide both methods with character joint positions $\mathbf{J}$.

\subsection{Experiment Results}
\textbf{Quantitative Comparison.} We report PMD and ELS metrics for ours and baseline methods in Table \ref{table:quant}. We observe traditional LBS-based methods \cite{li2021learning, liao2022skeleton} produce a significantly larger apparel error (larger PMD$_a$), as they fail to model dynamical effects on apparel. In addition, both methods often estimate inconsistent skinning weights on a single apparel instance, which causes deformed apparel to be torn apart with unnatural edge connections (lower ELS). In contrast, we explicitly model non-linear apparel deformation, thus generating superior apparel results. Moreover, thanks to the proposed geodesic attention, our method achieves an improved result in body deformation as well.

\begin{table}[htp!]
\centering
\caption{\textbf{Quantitative Comparison on the MMDMC test set.} Our method produce results with superior quality for both body and apparel parts compared to baseline methods \cite{li2021learning, liao2022skeleton}. } 
  \begin{tabular*}{.85\linewidth}{@{\extracolsep{\fill}}lcccc}
    \toprule
    \multicolumn{1}{l}{Methods} & PMD  ($cm$)$\downarrow$ & PMD$_a$ ($cm$) $\downarrow$ & PMD$_b$ ($cm$) $\downarrow$ & ELS $\uparrow$    \\ \hline
    \centering
  NBS \cite{li2021learning} & 4.63 & 8.90 & 1.84 & 0.73 \\
  SPT\cite{liao2022skeleton} & 2.53 & 6.62 & 0.83 & 0.81 \\ \hline
  Ours & \textbf{1.08} & \textbf{1.91} & \textbf{0.75} & \textbf{0.94} \\
  \bottomrule
  \end{tabular*}
  \label{table:quant}
\end{table}

\noindent
\textbf{Qualitative Comparison.} We show in Figure \ref{fig:quali} the qualitative comparison results. For characters with challenging complex topology, NBS fails to correctly estimate the body skinning weights, leading to twisted poses and discontinuous body parts. In addition, both methods do not produce realistic apparel animation, in which the apparel either does not locally deform or penetrates with the body. In contrast, we generate plausible and realistic apparel deformation, which improves the overall quality in motion transfer.

\begin{figure}[htp!]
{\includegraphics[width=\textwidth]{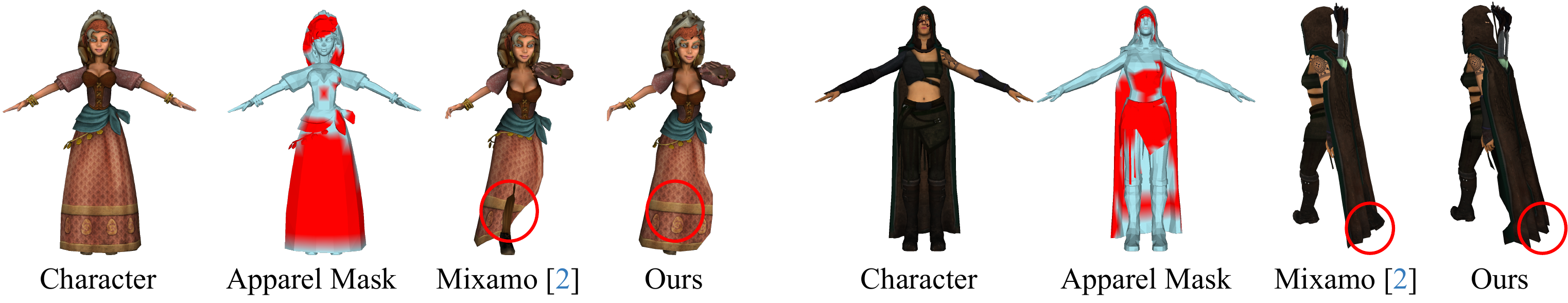}}
    \centering
    \caption{\textbf{Generalizability Evaluation. }The apparel animations in the Mixamo dataset \cite{Mixamo} contain artifacts of body-apparel penetration. In contrast, our pretrained model can infer plausible apparel mask and then improves the realism on the apparel.}
    \label{fig:gene}
\end{figure}

\noindent
\textbf{Generalizability Evaluation. }To validate the generalizability of the model and the efficacy of the pretraining on the proposed MMDMC dataset, in Figure \ref{fig:gene}, we compare with the apparel animations provided by the Mixamo dataset and results produced by our method. Since the Mixamo dataset does not provide rigging for apparel, its apparel animations often contain body-apparel penetration, which are not suitable for training and quantitative evaluation. In comparison, we evaluate our pretrained model on the selected characters and observe that it can infer plausible apparel masks and generate more realistic results. However, extending the pretrained priors on \emph{rigged} apparel animation to model real garment dynamics, \emph{e.g.} detailed wrinkles, remains challenging, and we encourage future works to explore more effective apparel priors for real human characters.

\noindent

\subsection{Ablation Study}

\begin{figure}[htp!]
{\includegraphics[width=\textwidth]{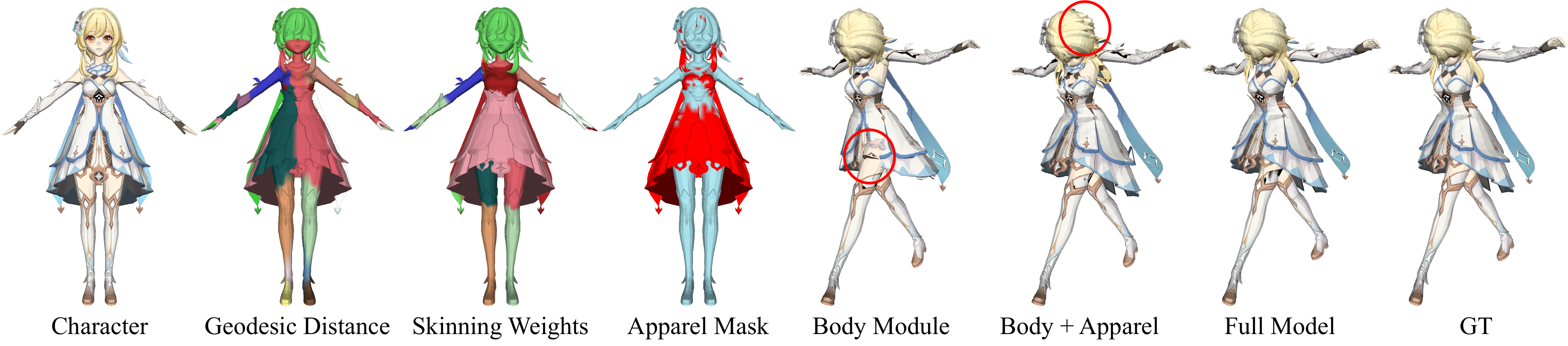}}
    \centering\caption{\textbf{Visualization of outputs. }We visualize outputs of intermediate modules and compare results from different design variants. The proposed geodesic attention block can effectively refine from noisy raw geodesic distance to estimate consistent skinning weights on complex character meshes, and the apparel segmentation network can estimate accurate apparel mask. We show using only the body module can generate unrealistic results with body-apparel penetration, while introducing the apparel module can refine apparel results, however, we observe discontinuity at body-apparel boundary. In comparison, the full model can effectively improve the overall quality. }
    \label{fig:ab1}
\end{figure}

\textbf{Effects of Each Module. }We show the effects of each module in Table \ref{table:ab1}.  
With explicit handling of the apparel deformation (Body + Apparel), we significantly reduce the apparel error compared to using only the body module (Body Module). However, we do not observe a significant improvement in the ELS score, as apparel vertices that are not correctly identified by the apparel segmentation module will remain deformed by the LBS method, which causes discontinuity. To address this issue, we further introduce the joint refinement module (Full Model) to encourage continuity, which achieves the best results. Finally, we show our method can be further improved if a ground truth apparel mask is provided (Full Model + GT Mask), which allows artists to freely adjust results with their own apparel annotations. We visualize intermediate results in Figure \ref{fig:ab1}.

\begin{table}[htp!]
\centering
\caption{\textbf{Effects of Proposed Modules.} We show our method (Full Model) achieves the best results compared to variants with only partial components or designs, and can be further improved with ground truth apparel masks.} 
  \begin{tabular*}{.85\linewidth}{@{\extracolsep{\fill}}lcccc}
    \toprule
    \multicolumn{1}{l}{Methods} & PMD  ($cm$)$\downarrow$ & PMD$_a$ ($cm$) $\downarrow$ & PMD$_b$ ($cm$) $\downarrow$ & ELS $\uparrow$    \\ \hline
    \centering
  Body Module & 2.37 & 5.42 & 0.78  & 0.82 \\
  Body + Apparel & 1.15 & 2.15 & 0.77 & 0.85 \\
  Full Model & \textbf{1.08} & \textbf{1.91} & \textbf{0.75} & \textbf{0.94} \\ \hline
  Full Model + GT Mask & 1.02 & 1.85 & 0.74 & 0.96 \\
  \bottomrule
  \end{tabular*}
  \label{table:ab1}
\end{table}

\noindent
\textbf{Effects of Geodesic Attention. } To show the effectiveness of the proposed geodesic attention block, we compare with two baselines: (\emph{i}) without using the geodesic distance (w/o $\mathbf{G}$) and (\emph{ii}) sorting joint positions based on the geodesic distances and concatenate them as vertex features (Sort by $\mathbf{G}$), following the design in \cite{RigNet}. For quantitatively comparison, we disable apparel physics in the ground truth data and show the result in Table \ref{table:ab2}. We observe that our method ($\mathbf{G}$ + Attention) achieves the best deformation result. We further visualize in Figure \ref{fig:ab1} and show that we can estimate smooth and consistent skinning weight conform to the body semantics, refined from the raw geodesic distance prior.

\begin{table}[htp!]
\centering
\caption{\textbf{Effects of Geodesic Attention.} We show the proposed geodesic attention ($\mathbf{G}$ + Attention) achieves superior results compared to other design choices.} 
  \begin{tabular*}{.85\linewidth}{@{\extracolsep{\fill}}lcccc}
    \toprule
    \multicolumn{1}{l}{Methods} & PMD  ($cm$)$\downarrow$ & PMD$_a$  ($cm$)$\downarrow$ & PMD$_b$  ($cm$)$\downarrow$ & ELS $\uparrow$    \\ \hline
    \centering
  w/o $\mathbf{G}$ & 1.35 & 3.17 & 0.74  & 0.84 \\
  Sort by $\mathbf{G}$ & 1.27 & 2.82 & 0.76  & 0.91 \\ \hline
  $\mathbf{G}$ + Attention & \textbf{1.14} & \textbf{2.29} & \textbf{0.69} & \textbf{0.93} \\
  \bottomrule
  \end{tabular*}
  \label{table:ab2}

\end{table}

\section{Discussion}

\textbf{Limitation \& Societal Impact.} Although our method achieves superior results, we rely on supervision from artists-designed rigging and physics properties of apparel, which requires substantial manual efforts. In addition, we only consider apparel on stylized humanoid characters with a unified skeleton, and have not modeled other types of characters, \emph{e.g.} quadruped animals. Furthermore, failure in deformation modules can cause broken body and apparel parts, or inappropriate dressing that is not suitable for public viewing.

\noindent
\textbf{Conclusion.} In this paper, we propose a novel method for high-quality motion transfer with realistic apparel animation. We create a new dataset MMDMC with detailed apparel annotations to facilitate the learning of apparel segmentation and deformation. Moreover, we introduce a geodesic attention block to incorporate semantic priors into the skeletal body deformation and devise an apparel deformation module to model the non-linear local deformation of apparel. Thanks to these efforts, our method effectively produces superior results on various characters and apparel.

\section*{Acknowledgements}
This research is funded in part by an ARC Discovery Grant DP220100800 on human body pose estimation and visual sign language recognition.

\bibliographystyle{splncs04}
\bibliography{main}
\end{document}